\documentclass[a4 paper, conference, final]{IEEEtran}
\usepackage{amsmath,amsfonts,amssymb,amsthm}
\usepackage{graphicx}
\usepackage{epsfig,epstopdf,url}

\usepackage{color}

\definecolor{az}{rgb}{0.6,0.0,0.0}
\definecolor{jc}{rgb}{0.0,0.6,0.0}
\usepackage{enumerate}

\graphicspath{{./perimmagine/}}

\begin{document}
	%
	\title{Revisiting Human Action Recognition: \\ Personalization vs. Generalization}

	\author{\IEEEauthorblockN{{Andrea Zunino}\IEEEauthorrefmark{1}\IEEEauthorrefmark{2}, {Jacopo Cavazza}\IEEEauthorrefmark{1}\IEEEauthorrefmark{2} and Vittorio Murino\IEEEauthorrefmark{1}\IEEEauthorrefmark{3}}
		\IEEEauthorblockN{\texttt{firstname.lastname@iit.it}}
		\IEEEauthorblockA{\IEEEauthorrefmark{1} Pattern Analysis \& Computer Vision, Istituto Italiano di Tecnologia, Via Morego 30, 16163, Genova, Italy}
		\IEEEauthorblockA{\IEEEauthorrefmark{2} Universit\`{a} degli Studi di Genova - Dipartimento di Ingegneria Navale, Elettrica, Elettronica e delle Telecomunicazioni, \\  Via All'Opera Pia, 11A, 16145, Genova, Italy}
		\IEEEauthorblockA{\IEEEauthorrefmark{3} Universit\`{a} di Verona - Dipartimento di Informatica, Strada le Grazie 15, 37134, Verona, Italy}
	}
	
	\maketitle
	
	
	\begin{abstract}
		%
		By thoroughly revisiting the classic human action recognition paradigm, this paper aims at proposing a new approach for the design of effective action classification systems. 
		Taking as testbed publicly available three-dimensional (MoCap) action/activity datasets, we analyzed and validated different training/testing strategies. In particular, considering that each human action in the datasets is performed several times by different subjects, we were able to precisely quantify the effect of \textit{inter-} and \textit{intra-subject variability}, so as to figure out 
		the impact of several learning approaches in terms of classification performance. The net result is that standard testing strategies consisting in cross-validating 
		the algorithm using typical splits of the data (holdout, k-fold, or one-subject-out) is always outperformed by a ``personalization'' strategy which learns
		how a subject is performing an action. In other words, it is advantageous to customize (\textit{i.e.}, personalize) the method to learn the actions carried out by each 
		subject, rather than trying to generalize the actions executions across subjects. Consequently, we finally propose an action recognition framework consisting 
		of a two-stage classification approach where, given a test action, the subject is first identified before the actual recognition of the action takes place. 
		Despite the basic, off-the-shelf descriptors and standard classifiers adopted, we noted a relevant increase in performance with respect to 
		standard state-of-the-art algorithms, so motivating the usage of personalized approaches for designing effective action recognition systems.
	\end{abstract}
	
	
	\IEEEpeerreviewmaketitle

	\section{Introduction}
	
	Classification of actions or activities\footnote{Action and activity are here used interchangeably, if not differently specified.} from videos or still images is actually a very complex task due to the high variability of the action/activity classes which can be performed differently by different people, and even differently by the same person at different times. Besides, other problems related to the context, clutter/noise, illumination variations, and occlusions make this task even more challenging. As a consequence, the two-dimensional (2D) and 2D+time actual informative structure of the image and video data is strongly affected by all the above issues, which complicate the design of reliable action recognition systems. 
	For all the aforementioned reasons and because of the implying relevant practical applications, activity recognition is undoubtedly one of the most interesting and debated problems in computer vision and pattern recognition \cite{poppe}. 
	Fortunately, this scenario slightly improves owing to recently introduced three-dimensional (3D) sensor technology, which can nowadays be used to safely and accurately capture human motion 
	at high-resolution 
	in both spatial and temporal domains (\textit{e.g.}, VICON), with good accuracy, and at low cost (\textit{e.g.}, Kinect). Despite the richer information provided by 3D+time data, the task is only apparently easier as, although some of the problems linked to the appearance may result mitigated, the fine 3D information which can be recovered by such data may degrade the performance of the adopted recognition method.
		
	\begin{figure}[t]
		\includegraphics[width=\columnwidth,keepaspectratio]{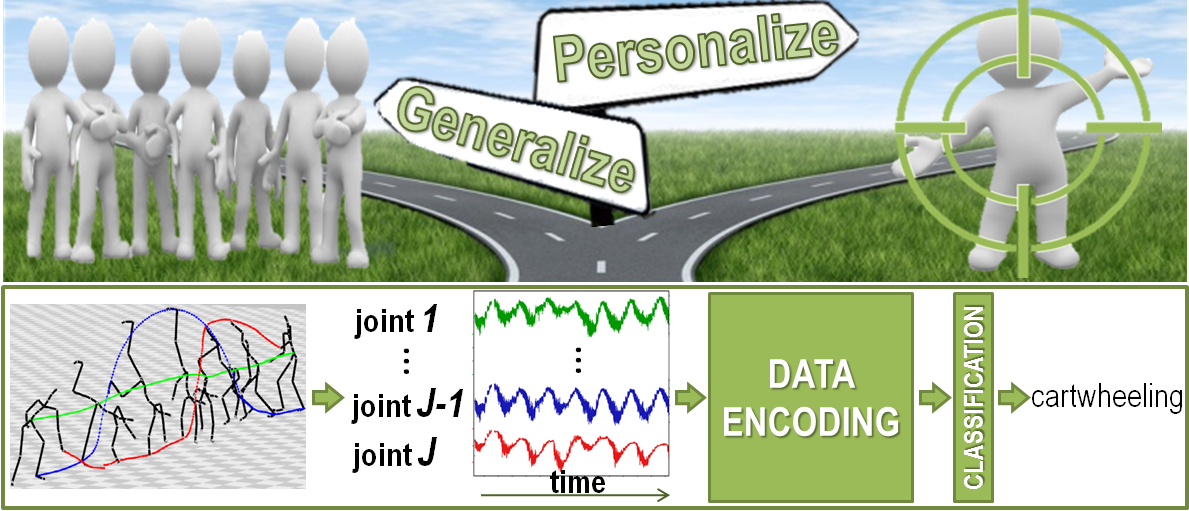}
		\caption{From MoCap data, action recognition systems perform automatic classification on 3D time series of joint positions (below). In this respect, we investigate the  counter-position between {\bf personalizing} on a single subject to boost the performance (top-right) against the {\bf generalizability} on unknown human agents, despite a decreased effectiveness (top-left).}
		\label{fig:top}
	\end{figure}
	All such aspects corroborate the continuous development and improvement of computational approaches for 3D action recognition, which can be categorized into three main classes, according to \cite{LoPresti}. In the first one, the whole set of joints coordinates is modeled, for instance, by using first/second order finite difference schemes \cite{movpose} or bag-of-word encodings \cite{Xia:12}. Second, selection criteria of the most discriminant joint can be devised; for instance, \cite{Ofli:13} exploited mean and variance of joint angles and maximal angular velocity. Third, a temporal modeling of dynamics is performed through different techniques (\textit{e.g.}, autoregressive models \cite{slama} or non-parametric Bayesian inference \cite{raman}). Recently, \cite{beyond} set the new state-of-the-art by considering a double layer of low- and high-level kernels to represent each action separately and to measure their mutual similarity, respectively.
	
	
	In such context, this work undertakes a revisiting perspective of the action recognition paradigm, probing the principal evaluation strategies applied in the literature on the most common, publicly available, benchmark datasets. Thus, we aim at providing a deep understanding about the challenges that have to be faced when devising classification protocols: such awareness leads us to introduce a new effective, yet simple, approach for action recognition.
	The experimental testbed we have chosen consists of 3 public datasets, namely MSR-Action3D \cite{Action3D}, MSRC-Kinect12 \cite{MSRC} and HDM-05 \cite{HDM-05}. Each has own peculiar traits, \textit{e.g.}, the amount and type of considered action classes or the number of skeletal joints. However, a common shared aspect is that a same action is performed by several subjects and a same subject actually performs each action more times. The variability of considered actions aim at reproducing real-world scenarios, while repeating actions and considering multiple actors allow to increase the learning methods in robustness and generalization, respectively. Usually, action recognition methods in the literature do not exploit the information associated to the subject identity, but they typically consider different splits of all action instances (\textit{e.g.}, k-fold cross-validation) in the training/testing phases. Nevertheless, such information is quite relevant, indeed discriminant, for the actual recognition of the actions since \emph{each} human being shows peculiar features which are reflected in the way an action is performed. The former aspects have been rarely investigated and seldom quantified by previous recognition system to date and, to this end, we focus on two main aspects:

	
	\begin{itemize}
		\item \textit{Inter-subject variability}, which either refers to anthropometric differences of body parts or to incongruous personal styles in accomplishing the scheduled action. In practice, different subjects may perform the same (even very simple) action in different ways.
		\item \textit{Intra-subject variability}, which represents the random nature of each single action class (\textit{e.g.}, throwing a ball), which can also be dictated by pathological conditions or environmental factors. In other words, this reflects the fact that a subject never performs an action in the same exact way.
	\end{itemize}
	
	Both aspects lead to the fact that a same action could not be performed exactly equal to itself, either it is executed by the same or different human beings. In this line, the additional information of subject identity has empirically demonstrated to be effective in customizing the classification on a specific user for speech \cite{speech}, handwriting \cite{hand}, and gesture \cite{VGool} recognition. 
	
	Among the few works which studied the variability within/across subjects, for instance, \cite{Bao:04} did not register a strong impact of different subjects in daily activities classification, and \cite{Dalton:13} documented the stability of the performance on an \textit{ad hoc} acquired dataset characterized by biometric homogeneity of the participants. Differently, in \cite{Taylor:10}, the performance of checking the correct execution of gymnastics sharply falls when the subject under testing is excluded from the training phase. A similar trend was registered by \cite{Tormene:09} and \cite{Yurtman:14} for computer assisted rehabilitation tasks, as well as by \cite{Barshan:15} which performed a theoretical dissertation about within-subject and across-subjects noise using wearable motion sensors. Globally, \cite{Bao:04,Dalton:13,Taylor:10,Tormene:09,Yurtman:14,Barshan:15}  did not mutually agree in their conclusions and, also, their investigation is actually limited by the use of private datasets explicitly designed for the considered application. 
	
	Despite some previous approaches grant in some way the importance of the knowledge of the human subject (especially for rehabilitation purposes, where the goal is directed to a specific subject), no study has been systematically reported to date on commonly used and publicly available datasets for general action/activity recognition. 
	In other words, it is still an open problem to quantify how much those datasets are affected by \textit{inter-} and \textit{intra-subject variability},
	%
and hence to figure out the impact of subjectiveness in action recognition to actually investigate the trade-off between personalization and generalization in the design of automatic systems.	
	\\ These arguments are investigated in this paper through the following main contributions.
	
	\begin{enumerate}[$(i)$]
		\item Considering MSR-Action3D, MSRC-Kinect12 and HDM-05 benchmark datasets, we analyse different testing strategies, investigating the cases where 1) the data of one subject are left out for testing -- \textbf{\textit{One-Subject-Out}}, 2) the actions (properly split in separate sets) executed from all subjects are present in both training and in testing -- \textbf{\textit{Cross-Validation}}, and 3) the classification is performed over a simplified problem considering only the instances belonging to one specific subject at a time -- \textbf{\textit{Personalization}}. Such analysis is performed considering two different types of off-the-shelf encodings: covariance-based feature description 
		and dynamic time warping, both used to estimate the mutual similarity of different action instances in the context of an SVM-based classification approach.
		\item The role of subjectiveness is introduced and investigated, to estimate the balance between personalization and generalization (Figure \ref{fig:top}). By means of a quantitative statistical analysis, we evaluate the effectiveness of retrieving in testing all the subjects used in the training phase by assessing the role played by either \textit{inter-} or \textit{intra-subject variability}.
		\item Finally, we propose a two-stage recognition pipeline where the preliminary identification of the subject is followed by a subject-specific action classification. Overall, our new proposed pipeline shows a strong performance with respect to both \textit{Cross-Validation} and \textit{One-Subject-Out} strategies, also being superior to the state-of-the-art method \cite{beyond}.
	    This promising result may open a new paradigm for the development of action/activity recognition systems, also embracing the possibility to exploit these findings in the design of joint biometric, kinematic-driven authentication systems. 
	\end{enumerate}
	
	The rest of the paper is organized as follows. In Section \ref{sez:dat}, we present the considered datasets and the feature representations adopted, and the evaluation strategies investigated are reported in \ref{sez:testmod}. Section \ref{sez:res} presents and widely discusses the experimental results, and we illustrate the aforementioned two-stage classification pipeline in Section \ref{sez:aut}. Finally, Section \ref{sez:conc} draws the conclusions of this study.

\section{Datasets \& Feature Encoding}\label{sez:dat}

Our investigation involves three publicly available MoCap datasets for activity recognition: MSR-Action3D, MSRC-Kinect12 and HDM-05. For all our experiments, we only used the 3D skeleton coordinates while the other data available (\emph{e.g.}, depth maps or RGB videos) were not taken into account. For the sake of clarity we briefly introduce each of them.

$\boldsymbol{-}$ The {\bf MSR-Action3D} \cite{Action3D} dataset has 20 action classes of mostly sport-related actions (\emph{e.g.}, \emph{jogging} or \emph{tennis-serve}), performed by 10 subjects. $J=20$ joints are extracted from the Kinect sensor data to model the human pose of the human agents. Each subject performs each action 2 or 3 times. In total, we used 544 sequences, once removed the more corrupted ones \cite{egizi}. 
 
$\boldsymbol{-}$ {\bf MSRC-Kinect12} \cite{MSRC} is a relatively large dataset of 3D skeleton data, recorded by means of a Kinect sensor. The dataset has 594 sequences, containing 12 action classes performed by 30 different subjects, precisely there are 6244 annotated gesture instances in total. Each subject accomplishes each class of action 16 times, on average. The available motion files contain the trajectories estimated for $J=20$ 3D skeleton joints. 

$\boldsymbol{-}$ In {\bf HDM-05} \cite{HDM-05}, the number of skeleton joints is $J=31$, and the dataset contains more than three hours of systematically recorded VICON MoCap data acquiring different types of gestures performed by 5 professional actors. Motion clips have been manually cut out and annotated into roughly 100 different motion classes. In this work we have removed some severely corrupted samples (as in \cite{egizi}) and, as in \cite{beyond}, we selected only the following 14 classes:  \emph{clap above head}, \emph{deposit floor}, \emph{elbow to knee}, \emph{grab high}, \emph{hop both legs}, \emph{jog}, \emph{kick forward}, \emph{lie down floor}, \emph{rotate both arms backward}, \emph{sit down chair}, \emph{sneak}, \emph{squat}, \emph{stand up lie} and \emph{throw basketball}. In this setup each subject accomplishes each action 5 times, on average.

For all the aforementioned datasets, each trial can be formalized as a collection $\mathbf{S}$ of $\tau$ different acquisitions $\mathbf{p}(1),\dots,\mathbf{p(\tau)}.$ For any $t = 1,\dots,\tau,$ we denote with $\mathbf{p}(t)$ the column vector which stacks $\mathbf{p}_1(t),\dots,\mathbf{p}_{J}(t) \in \mathbb{R}^3$, the three-dimensional $x,y,z$ coordinates of the $J$ skeletal joints. Using this notation, we now briefly introduce the two different representations for MoCap data.

First, we investigated the usage of dynamic time warping ($\mathrm{DTW}$), a classical tool to quantify the similarity across two different time series by means of alignment \cite{DTW,negDTW}. In order to apply $\mathrm{DTW}$, we evaluated the differences between any two joints collection $\mathbf{S} = [\mathbf{p}(1),\dots,\mathbf{p}(\tau)]$ and $\mathbf{S}' = [\mathbf{p}'(1),\dots,\mathbf{p}'(\tau')]$ through the following distance
\begin{equation}\label{eq:ff}
d(\mathbf{p}(s), \mathbf{p}'(t)) = \frac{1}{J} \sideset{}{_{j=1}^{J}}\sum {\|\mathbf{p}_j(s) - \mathbf{p}'_j(t)\|},
\end{equation}
where $\|\cdot\|$ is the Euclidean norm, $s = 1,\dots,\tau$ and $t = 1,\dots,\tau'$. The final
similarity measure, provided by $\mathrm{DTW}$ to compare $\mathbf{S}$ and $\mathbf{S}'$, is $\delta(\mathbf{S},\mathbf{S}')$ which is the minimum value of \eqref{eq:ff} computed over all the sequences of timestamps which optimally align $\mathbf{S}$ with $\mathbf{S}'$ (see \cite{DTW} for more details).

Second, we also estimated the $n \times n$ covariance matrix
\begin{equation}\label{eq:cov}
\mathcal{C} = \dfrac{1}{\tau -1} \sum_{t = 1}^\tau (\mathbf{p}(t) - \overline{\mathbf{p}})(\mathbf{p}(t) - \overline{\mathbf{p}})^\top,
\end{equation}
related to any trial $\mathbf{S}$, where $\overline{\mathbf{p}} = \frac{1}{\tau} \sum_{s = 1}^\tau \mathbf{p}(s)$ averages all the $\tau$ coordinates and we denote $n = 3J$ for convenience. Since $\mathcal{C}$ is positive definite, we thus exploited the theory of the Riemannian manifold $Sym^+_n$ and projected \eqref{eq:cov} onto the tangent space to obtain $\widetilde{\mathcal{C}}$ \cite{Jayasumana:13}. Then, using the symmetry of $\widetilde{\mathcal{C}},$ we extracted its independent entries, yielding the following $n(n+1)/2$ vector
\begin{equation}
\mathrm{COV} = [\widetilde{\mathcal{C}}_{11},\dots,\widetilde{\mathcal{C}}_{1n},\widetilde{\mathcal{C}}_{21},\dots,\widetilde{\mathcal{C}}_{1n},\dots,\widetilde{\mathcal{C}}_{nn}].
\end{equation}
Note that the usage of covariance is inspired by \cite{beyond}, which set the new state-of-the-art performance for action recognition from MoCap data. Also, our approach is similar to the case $L=1$ in \cite{egizi}, where a $L$-layered temporal hierarchy of covariance descriptors is proposed, but differently from us, the projection stage onto the tangent space is not considered.

For both representations, we used the support vector machine\footnote{In all experiments, for the SVM cost parameter, we fixed $C=10$.} (SVM) for classification: when fed with $\mathrm{COV}$, we normalized the data imposing zero mean and unit variance and we then used a linear kernel. Instead, the negative dynamic time warping kernel function \cite{negDTW} produced the training and testing Gram matrices given in input to the SVM.

\section{Evaluation Strategies}\label{sez:testmod}
As previously noticed, the common aspect across the considered datasets is that each subject performed the same action class several times (\textit{e.g.}, 2-3 times in the {MSR-Action3D}). Hence, we want to check if the additional knowledge of the identity of subject who is acting can boost the classification. Thus, we introduce the three different testing modalities adopted.  

\textbf{\textit{One-Subject-Out}} considers as testing data all the action instances belonging to one subject only, while the remaining trials are used in training. Consequently, the final classification results average all the subject-related intermediate scores. This is the more appropriate procedure in terms of generalization. For instance, it is fundamental for real-time 24h/7d video-surveillance application where the system should be able to recognize activities performed by never seen human agents. Moreover, \textit{One-Subject-Out} is in line with the cross-subject test setting adopted in \cite{beyond,Action3D}. 

In the \textbf{\textit{Cross-Validation}} strategy we collect the data coming from any subject in a way that, for each of them, $\frac{2}{3}$ of samples are used in training and the remaining $\frac{1}{3}$ in testing. To guarantee robustness, the final classification results are averaged over 20 random choices for such a partition of the data. Such procedure can boost the action recognition accuracy since, during training, it exploits the information regarding all the subjects: in this way, the classifier can more easily discriminates the action performed by a subject on which has been already trained\footnote{Please note that a test sample is never seen by the system in training.}.

In the \textbf{\textit{Personalization}} strategy, we suppose to have the same number of classifiers as the number of the subjects present in each dataset. Each classification model is specific of a given subject and it is trained only on his/her activity instances: to do this, once the classes with a number of trials less than $2$ are removed, we randomly choose $\frac{2}{3}$ of samples referring to every action. Thus, we test on the remaining $\frac{1}{3}$. The final accuracy score compares all the predicted and true labels, fusing the results of all the specific classifiers. As previously done, we average the classification results over 20 random splits of all the subject-specific instances.

\section{Experimental Results}\label{sez:res}
In this Section, we compare \textit{One-Subject-Out}, \textit{Cross-Validation} and \textit{Personalization} by means of the two types of features previously introduced in Section \ref{sez:dat}. Table \ref{tab:dtw} and Table \ref{tab:cov} report the results related to $\mathrm{DTW}$ and $\mathrm{COV}$, respectively. 

In the majority of the comparisons, the $\mathrm{COV}$ descriptor obtains higher performance with respect to $\mathrm{DTW}$. Nevertheless, we can observe a common trend: the action classification performance grows when switching from \textit{One-Subject-Out} to \textit{Cross-Validation}, reaching its peak with \textit{Personalization}. Moreover, such behavior is also independent from the features adopted since showed by both $\mathrm{DTW}$ and $\mathrm{COV}$. 
Additionally, it is worth noting that the results reported in Tables \ref{tab:dtw} and \ref{tab:cov} show that the accuracies obtained with the three different modalities are inversely proportional to the number of the samples used in the training phase. In all the cases, the lowest performance is always scored adopting \textit{One-Subject-Out}, although it has the largest number of training samples. This result is expected, since \textit{One-Subject-Out} has to extrapolate more regular patterns from the data, namely finding subject-invariant characteristics to recognize actions of unseen human agents.
 In other words, in \textit{One-Subject-Out} procedure, SVM does not learn how a specific subject performs an action but how a particular action is generally fulfilled. Conversely the \textit{Personalization} strategy obtains the best results for all datasets. This procedure considers in the training step the least number of samples, only regarding a specific subject. In particular, if we focus on the acquisition details of MSR-Action3D dataset (see Section \ref{sez:dat}), very few trials per activity performed by the same subject are available (sometimes only one action instance per subject is taken for training). Despite this, \textit{Personalization} scores $92.46\%$ and $81.75\%$ with $\mathrm{COV}$ and $\mathrm{DTW}$ features respectively, and outperforms all the other two strategies. Indeed, the other two datasets, MSRC-Kinect12 and HDM-05, are almost saturated using both descriptors (\textit{e.g.}, $99.57 \pm 0.16$ of $\mathrm{DTW}$ on MSRC-Kinect12 and $99.02 \pm 0.98$ of $\mathrm{COV}$ on HDM-05).
 
\textit{Cross-Validation} results deserve an own discussion since, in all the comparisons, it always gives intermediate classification scores. This is a hybrid strategy with respect to the previous ones since exploiting information coming from the totality of the available subjects but does not specialize the discriminative model over a specific actor. All the scores reported are always lower than the \textit{Personalization} accuracies although, in some cases, the gap between them is very small (\textit{e.g.}, \textit{Cross-Validation} scores about 1\% less with respect to \textit{Personalization} on MSRC-Kinect12 dataset, see Table \ref{tab:cov}). Actually, adding many training samples belonging to different subjects does not always lead to an improvement, but, frequently, the SVM classifier gets confused. Indeed, if we compare the three modalities, the performance of \textit{Cross-Validation} is higher than \textit{One-Subject-Out} one since, in the former, the information of the tested subject is likely available also in the training phase. At the same time, it is evident that, with \textit{Personalization}, a classifier can be trained using much less training data but still extracting enough discriminative patterns. Conversely, it seems that \textit{Cross-Validation} does not benefit so much from the augmented amount of training data. In the following Section, we provide a better understanding about this phenomena.  

 \begin{table}[h!]
 	\centering
 	\caption{$\mathrm{DTW}$ classification accuracies on the three MoCap datasets. Mean and standard deviation are reported in percentages for each testing strategies (best results are in bold).} 	
 	\includegraphics[width=\columnwidth,keepaspectratio]{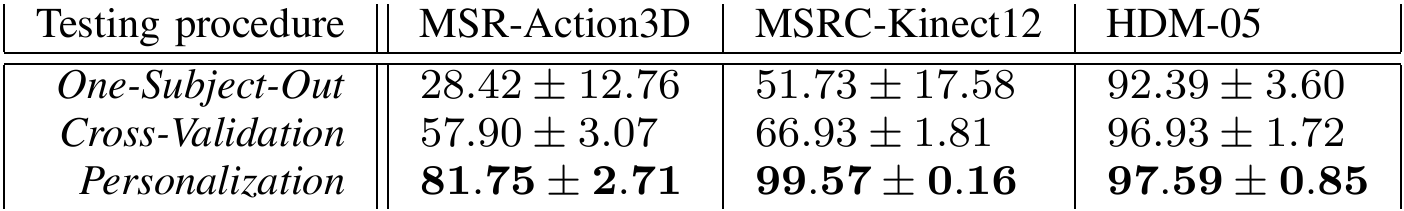}
 	\label{tab:dtw} 
 	\vspace{-.1 cm}
 \end{table}
 
 \begin{table}[h!]
 	\centering
 	\caption{$\mathrm{COV}$ classification accuracies on the three MoCap datasets. Mean and standard deviation are reported in percentages for each testing strategies (best results are in bold).} 	
 	\includegraphics[width=\columnwidth,keepaspectratio]{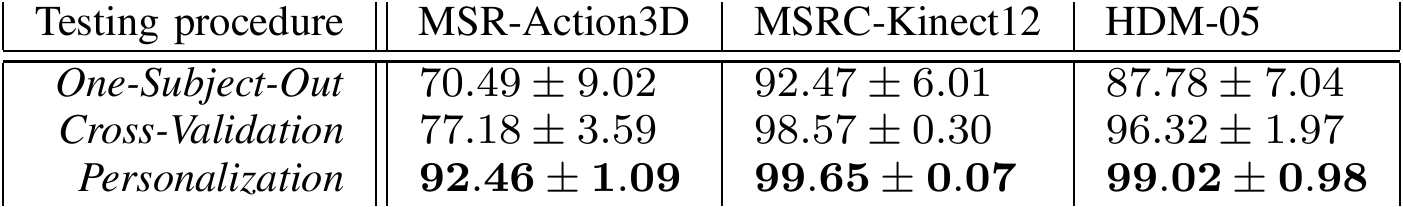}
 	\label{tab:cov} 	
 \end{table}

\subsection{Quantitative statistical analysis} 

To provide a better understanding about personalization vs. generalization and \textit{inter-} \& \textit{intra-subject variability} on the considered datasets, we report some quantitative results obtained by means of the following statistics.

$\bullet$ Once fixed the \textit{Cross-Validation} strategy, we aim at checking which source of information is at disposal of SVM when an exact prediction is performed. Hence, we compute the probability $p_\mathrm{subject}$ that, on average, any correctly classified testing sample and the training sample closest to it both belong to the same subject. Clearly, high/low $p_\mathrm{subject}$ values check if testing on the same subjects used for training gives a pros/cons for the classification, respectively.

$\bullet$ For each action class, if considering one trial and its nearest in the feature space, we compute the probability that, on average, they belong to the same subject. We call it $p_\mathrm{inter}$ since absolutely quantifying \textit{inter-subject variability}, whose impact is negligible if $p_\mathrm{inter} \approx 0$.

$\bullet$ To do the same with the \textit{intra-subject variability}, we compute $p_\mathrm{intra}$ by checking whether, on average, for any fixed subject, every two samples, which are the closest ones in the feature space, actually refer to different classes of actions. From the definition, if $p_\mathrm{intra} = 0,$ all the trials of a given action and a given subject are almost identical and \textit{intra-subject variability} is totally absent. 

\begin{figure}[t!]
	\vspace{.2 cm}
	\centering
	\includegraphics[width=.45\columnwidth,keepaspectratio]{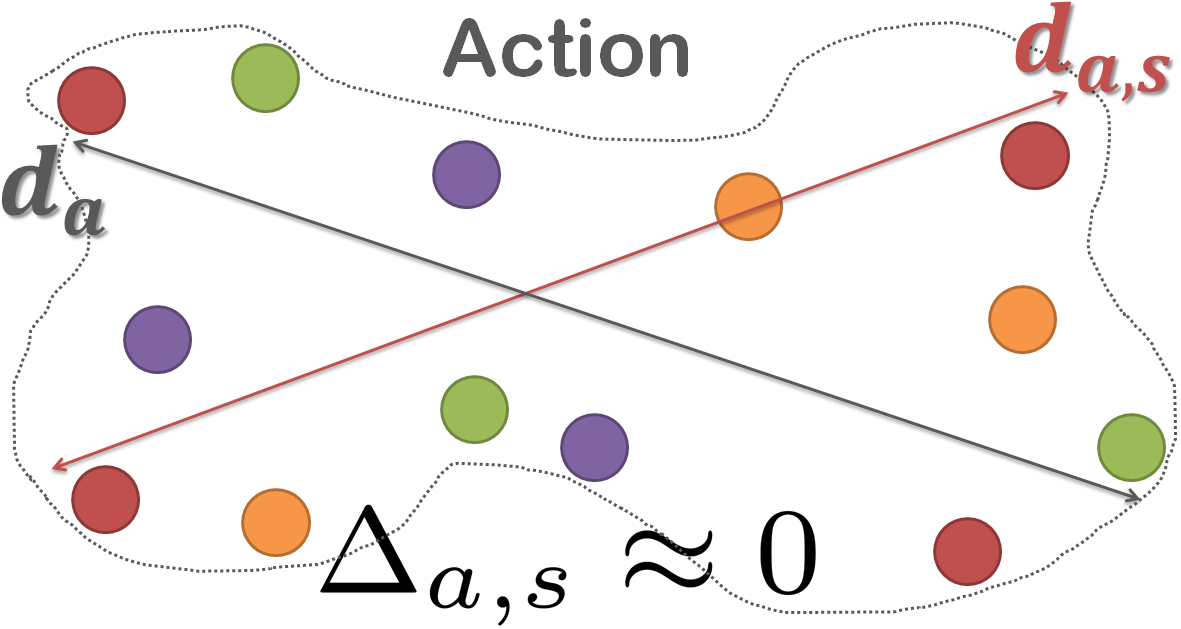}\qquad
	\includegraphics[width=.45\columnwidth,keepaspectratio]{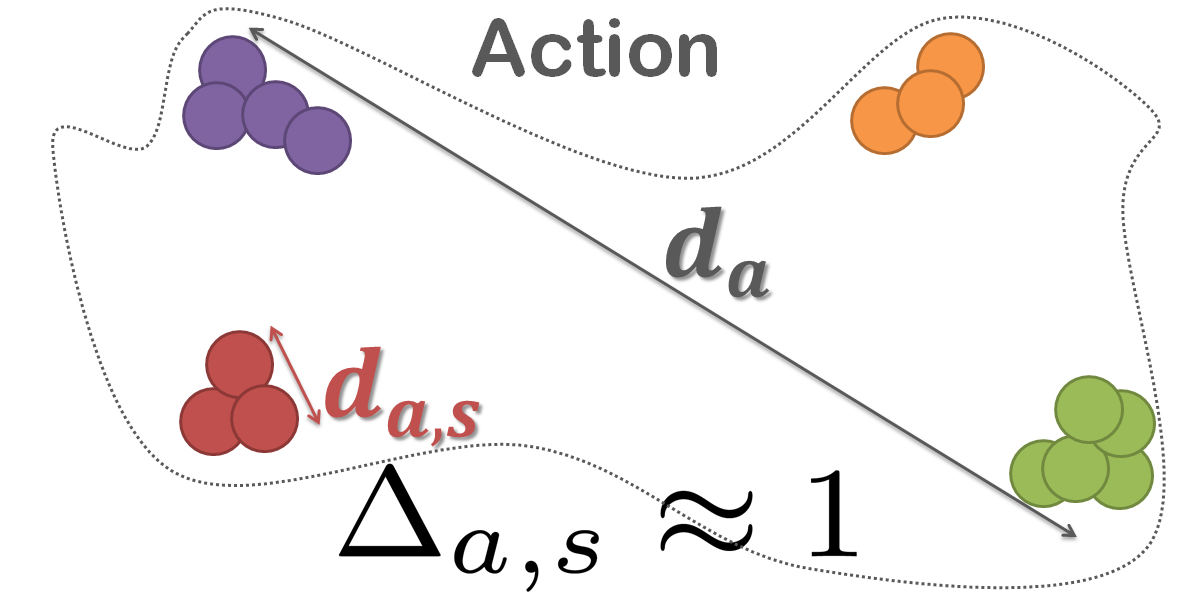}
	\caption{In the feature space, we surround the region referring to a single action. Within, each point represents a trial and different colors relate to different subjects. Left: When $\Delta_{a,s} \approx 0$, \textit{inter-subject variability} is minimized since, in general, trials from different subjects occupy nearby positions. Right: The case $\Delta_{a,s} \approx 1$ minimizes the \textit{intra-subject variability} because all the instances of the same subject are compactly clustered.}
	\label{fig:delta}
\end{figure}

$\bullet$ Finally, we devise a comprehensive metric for \textit{inter} and \textit{intra-subject variability}. Considering each action class $a$ separately, we find the two elements of maximal mutual distance, namely $d_a$. Also, we compute the maximal distance $d_{a,s}$ between all trials of action $a$ performed by subject $s.$ Then, once set $\Delta_{a,s} = \frac{|d_{a,s} - d_a|}{d_a}$, this metric spans the extremal cases $\Delta_{a,s} = 0$ and $\Delta_{a,s} = 1$. If $\Delta_{a,s} = 0$, then $d_{a,s} = d_a$ and the \textit{inter-subject variability} is minimized: instances belonging to the same action but different subjects are generally close to each others. Conversely, when $\Delta_{a,s} = 1,$ it implies $d_{a,s} = 0$ which minimizes the \textit{intra-subject variability} since all the trials of action $a$ and subject $s$ collapse into a single point. Clearly, it means that subject $s$ performs action $a$ in the same manner across all the different trials. We name $\Delta$ the final statistic, averaging $\Delta_{a,s}$ over $a$ and $s$. Globally, $\Delta$ measures which out of \textit{inter} or \textit{intra-subject variability} is overwhelming (see Figure \ref{fig:delta} for a better understanding).

Referring to all the aforementioned statistics $p_{\rm subject}$, $p_{\rm inter}$, $p_{\rm intra}$ and $\Delta$, a notion of ``closeness'' is involved. Clearly, it depends to the type of used feature: for $\mathrm{COV},$ the distance is the Euclidean one, since induced by a linear kernel. Instead, for DTW, we use the dynamic time warping distance $\delta$, as introduced in Section \ref{sez:dat}. Table \ref{tab:scleri} reports the values of all the previous statistics for all the considered datasets. We only report the values related to $\mathrm{COV}$ since no remarkable differences are registered when moving to $\mathrm{DTW}$\footnote{For instance, the value of $p_{\rm subject}$ for MSR-Actio3D is 0.77, for MSRC-Kinect12 is 0.97 and for HDM-05 is 0.85.}. Table \ref{tab:scleri} gives us useful insights about the classification results reported in Tables \ref{tab:dtw} and \ref{tab:cov}. All the datasets show a similar behavior.

Actually, in all cases, $p_{\rm subject}$ is extremely high (\textit{e.g.}, 0.89 for HDM-05) and
it indicates that each subject performs any action according to his/her own peculiarities which are effective when discriminating on the same subject. However, adding the information referring to other subjects can be misleading and this is motivated by the fact that \textit{Personalization} scores superior to \textit{Cross-Validation}: \textit{e.g.}, using $\mathrm{COV}$ on HDM-05, $99.02\%$ for the former and $96.32\%$ for the latter (Table \ref{tab:cov}).


Differently, the high values of $p_{\rm inter}$ certify that, in all the datasets, the \textit{inter-subject variability} is high and it frequently happens that different subjects have their own fashion to perform a same action, so classifying a given action of an unknown subject can therefore be difficult. As a consequence, this explains why, sometimes, \textit{One-Subject-Out} gives very poor performance (\textit{e.g.}, Table \ref{tab:dtw}, $\mathrm{DTW}$ on MSR-Action3D). 

The scored valued for $p_{\rm intra}$ attests that, in all the considered datasets, the \textit{intra-subject variability} is totally not problematic and, almost surely, each subject repeats a single action coherently along his/her trials. Also, such trend is confirmed by the excellent performance scored by \textit{Personalization}.

\begin{table}[t!]
	\centering
	\caption{Quantitative evaluation of \textit{inter} and \textit{intra-subject variability}}.
	\label{tab:scleri}
	\begin{tabular}{|r||c|c|c|c|}
		Dataset & $p_{\rm subject}$ & $p_{\rm inter}$ & $p_{\rm intra}$ & $\Delta$ \\ \hline \hline
		MSR-Action3D & 0.78 & 0.86 & 0.19 & 0.71 \\
		MSRC-Kinect12 & 0.97 & 0.97 & 0.01 & 0.90 \\
		HDM-05 & 0.89 & 0.95 & 0.01 & 0.74
	\end{tabular}
\end{table}

In conclusion, the registered values for $\Delta$ are, in all the cases, quite close to $1$. It means that all the instances of a given subject performing a single action occupy a compact region in the feature space. This is an opposed situation to a sparser configuration where the subjects are extremely shuffled. In other words, for all the datasets, different subjects can be considered as different sub-problems: \textit{Personalization} applies the \textit{divide et impera} principle, separately solving each of them. Conversely, \textit{One-Subject-Out} looks for patterns which have to be, at the same time, subject-invariant and action-specific. Despite the latter approach ensures much more generalization, it is also much more harder than the former: this is why \textit{Personalization} outperforms \textit{One-Subject-Out}.


\section{Two-stage Recognition Pipeline}\label{sez:aut}
In all the experiments related to \textit{Personalization}, we assume that the subject who is performing the given action is known a priori, so that we can easily choose the correct classification model trained over the specific human agent. In real-world applications, \textit{Personalization} can be replaced with a two-stage process where $1)$ a unique SVM model (\textit{subject-SVM}) classifies the identity of the subject and
$2)$ the final stage of action recognition is performed by means of a set of subject-specific SVM classifiers (\textit{action-SVMs}).
Both \textit{subject-SVM} and \textit{action-SVMs} are trained on the same random partition of the data (collecting $2/3$ of the trial from any subject and action). Actually, for \textit{subject-SVM}, no preliminary subject-dependent splitting is required, while, the trials belonging to a single subject at a time are used in the case of \textit{action-SVMs}.
In the testing phase, the \textit{subject-SVM} predicts the agent label that indexes which \textit{action-SVMs} classifier has to be used for the final action or activity recognition: when a subject is misclassified, the wrong model is used and, obviously, the action classification may be wrong. 

 \begin{table}[h!]
 	\centering
 	\caption{Two-stage recognition pipeline accuracies.}
 	 	\begin{tabular}{|r||l|l|l|}
 	 		 & {MSR-Action3D} & {MSRC-Kinect12} & {HDM-05}\\\hline\hline
 	 		\textit{subject-SVM}  & $90.74 \pm 2.41$ & $85.18 \pm 0.55$ & $85.67 \pm 3.18$ \\
 	 		\textit{action-SVMs}  & $\mathbf{90.46 \pm 1.17}$ & $\mathbf{97.14 \pm 0.39}$ & $\mathbf{97.03 \pm 1.36}$ \\ 		
 	 	\end{tabular}
 	\label{tab:twostep} 	
 \end{table}

 To validate our proposed pipeline, both \textit{subject-SVM} and \textit{action-SVMs} are fed with $\mathrm{COV}$ features, more powerful than $\mathrm{DTW}$. The results in Table \ref{tab:twostep} provide the mean and standard deviation of the accuracies scored in the two steps separately, over 20 different random partitions of the data. Since $\mathrm{COV}$ were originally suited for action recognition, such descriptors are not optimal for human identification. In fact, in all the datasets, the average accuracy of \textit{subject-SVM} (second row in Table \ref{tab:twostep}) does not exceed $91\%$
and, clearly, more suitable biometric descriptors could improve it\footnote{In the case of a perfect human identification provided by \textit{subject-SVM}, the action classification accuracy of \textit{action-SVMs} will match the \textit{Personalization}.}. Finally the \textit{action-SVMs} results (in bold in Table \ref{tab:twostep}) small deviate from the classification scores of \textit{Cross-Validation} and \textit{Personalization} (Table \ref{tab:cov}): \textit{e.g.} in MSRC-Kinect12, the action recognition score of \textit{Personalization} and the two-stage pipeline is $99.65 \pm 0.07$ and $97.14 \pm 0.39$, respectively. As a final remark, it is notable that, despite the off-the-shelf feature employed, the mean scored performance in action classification of the two-stage pipeline is superior to the state-of-the-art method \cite{beyond} on MSRC-Kinect12 ($92.3\%$) and HDM05 ($96.8\%$).
\section{Conclusions}\label{sez:conc}

In this paper, we investigated the generalization capability of automatic action and activity recognition systems focusing on \textit{Personalization} in a comparison with standard \textit{Cross-Validation} and \textit{One-Subject-Out} strategies. To this aim, we exploited $\mathrm{DTW}$ and $\mathrm{COV}$ on MSR-Action3D, MSRC-Kinect12 and HDM-05 benchmark MoCap datasets. 

From the experiments, \textit{One-Subject-Out} resulted the more challenging strategy, although being able to ensure a major generalizable. Differently, despite \textit{Cross-Validation} was actually boosted from the usage of the same subject in both training and testing, the additional information relative to the other subjects could mislead.
Finally, the \textit{Personalization} strategy, gave the highest performance, despite the lowest number of instances used in training. 

In addition, we also provided several quantitative statistics to measure \textit{inter} and \textit{intra-class variability} on the considered datasets: as a result, the latter is almost marginal, while the former is the actual burden that has to be tackled when devising new techniques.

Finally, we proposed a two-step classification pipeline by first identifying the subject and second using subject-specific classifiers for the actual action recognition. This promising result may pave the way of a new paradigm for the design of action/activity recognition systems, also embracing the possibility to exploit these findings for custom human-specific action recognition systems.


\bibliographystyle{IEEEtran}
 \bibliography{fontiJC,fontiAZ}

\end{document}